\documentclass[letterpaper, 10 pt, conference]{ieeeconf}  % Comment this line out if you need a4paper

\IEEEoverridecommandlockouts                              % This command is only needed if 
\title{\LARGE \bf
GFS-VO: Grid-based Fast and Structural Visual Odometry
}

\author{Zhihe Zhang}% <-this % stops a space

\usepackage{algorithmicx}
\usepackage{algorithm}
\usepackage{amssymb}
\usepackage{amsmath}
\usepackage{algpseudocode}
\usepackage{booktabs}
\usepackage[table,xcdraw]{xcolor}
\usepackage{graphicx}
\usepackage{stfloats}
\usepackage{threeparttable}
\usepackage{cite}
\usepackage{multirow}
\begin{document}
\maketitle
\thispagestyle{empty}
\pagestyle{empty}

%%%%%%%%%%%%%%%%%%%%%%%%%%%%%%%%%%%%%%%%%%%%%%%%%%%%%%%%%%%%%%%%%%%%%%%%%%%%%%%%
\begin{abstract}
  In the field of Simultaneous Localization and Mapping (SLAM), researchers have always pursued better performance in terms of accuracy and time cost. Traditional algorithms typically rely on fundamental geometric elements in images to establish connections between frames. However, these elements suffer from disadvantages such as uneven distribution and slow extraction. In addition, geometry elements like lines have not been fully utilized in the process of pose estimation. To address these challenges, we propose GFS-VO, a grid-based RGB-D visual odometry algorithm that maximizes the utilization of both point and line features. Our algorithm incorporates fast line extraction and a stable line homogenization scheme to improve feature processing. To fully leverage hidden elements in the scene, we introduce Manhattan Axes (MA) to provide constraints between local map and current frame. Additionally, we have designed an algorithm based on breadth-first search for extracting plane normal vectors. To evaluate the performance of GFS-VO, we conducted extensive experiments. The results demonstrate that our proposed algorithm exhibits significant improvements in both time cost and accuracy compared to existing approaches.
\end{abstract}

%%%%%%%%%%%%%%%%%%%%%%%%%%%%%%%%%%%%%%%%%%%%%%%%%%%%%%%%%%%%%%%%%%%%%%%%%%%%%%%%
\section{INTRODUCTION}
\label{INTRODUCTION}

Simultaneous Localization and Mapping (SLAM) is a vital task in computer vision, enabling autonomous systems like robots, drones, and unmanned vehicles to navigate and create maps of unknown environments. A comprehensive SLAM framework typically consists of three core components: front end, back end, and loop detection. However, some frameworks opt to exclude loop detection to meet real-time and lightweight requirements. These frameworks are commonly known as odometry. Visual odometry is one such technique that has garnered substantial attention from the fields of computer vision and robotics. It utilizes sequences of images as input, providing advantages such as portability, cost-effectiveness, and robustness to environmental conditions.

In feature-based visual odometry, the utilization of geometry features plays a critical role in establishing frame-to-frame connections. Point features are commonly used due to their ease of extraction and abundance in the environment. However, they are susceptible to lighting variations, occlusion, and blur, resulting in decrease in pose estimation accuracy. One solution is incorporating line features into framework. Line features exhibit greater robustness to environmental factors compared to point features, offering more stable constraints between frames. However, existing research on line features has certain shortcomings, which can be summarized as follows:
\begin{enumerate}
  \item High cost of extraction. Existing approaches commonly employ LSD \cite{von2012lsd} as line extractor, which is readily accessible through OpenCV functions. However, the computational time required to calculate line-support region is prohibitively expensive, which contradicts the real-time requirements of visual odometry. 
  \item Inhomogeneous distribution of lines in the image. Both point and line features exhibit a common weakness of uneven distribution, being abundant in textured areas but scarce in regions with low texture. This imbalance frequently leads to pose estimation inaccuracies.  
  \item Underutilization of Line feature. During the process of pose estimation and optimization, line do not exhibit significant difference compared to point.
\end{enumerate}

In light of the aforementioned deficiencies, we present GFS-VO, a novel RGB-D camera-based visual odometry approach. Our contributions are outlined as follows:

\begin{itemize}
  \item We optimize extraction of line and analyze difficulties associated with line homogenization. To address these challenges, we propose three strategies that effectively achieve line homogenization.
  \item We design a plane normal vector extraction algorithm based on breadth-first search, which achieves faster and more accurate extraction of MA than existing methods.  
  \item We introduce a visual odometry framework that combines point and line features. A variety of constraints are employed to obtain more precise estimations of pose.
\end{itemize}

In the rest of this paper, we first provide an overview of related approaches in Sec. \ref{RELATED WORK}, then explain the details of our proposed framework in Sec. \ref{METHOD}, followed by experiments in Sec. \ref{EXPERIMENT RESULT} and expectation in Sec. \ref{CONCLUSION}.

\section{RELATED WORK}
\label{RELATED WORK}

In visual odometry, geometry features are widely used for pose estimation. Among these features, point features, as the most basic geometry element, play an essential role in various algorithms. They can be extracted and described quickly and accurately in most scenes, resulting in preferable performance of point feature-based frameworks \cite{mur2015orb, mur2017orb, campos2021orb, qin2018vins}. However, the sensitivity and instability of point features have prompted researchers to integrate more robust feature such as line feature into their framework \cite{pumarola2017pl, fu2020pl, he2018pl}.

Compared to point, line extraction is more time-consuming. The Line Segment Detector (LSD) \cite{von2012lsd} estimates the rectangular approximation of line support region based on the angle of each pixel and then calculate parameter of line. Although the gradient-based growing process is fast, the calculation and validation of support region incur time costs, ultimately affecting overall speed. \cite{akinlar2011edlines} also utilizes a growing process to detect lines but replaces the support region with anchors, which significantly accelerates extraction speed.

In addition to extraction speed, optimizations are employed to handle unique properties of line features. \cite{suarez2022elsed} address line cracks by connecting lines based on the current gradient direction. \cite{zhou2022edplvo}\cite{9484792} utilize collinear constraints to compensate for instability caused by fractures. Furthermore, the inhomogeneous distribution of geometry features also poses challenge. The quadtree structure adopted in \cite{mur2017orb} achieves point homogenization by placing feature points into nodes and preferentially dividing nodes based on the number of feature points within them. However, applying the same division approach to line features presents difficulties in allocating lines effectively, as incomplete grid coverage would compromise the homogenization effect.

Lastly, the utilization of line features in visual odometry frameworks remains limited. \cite{pumarola2017pl}\cite{9393474} calculate line reprojection using endpoint-to-line distance and minimize sum of point and line errors to estimate camera's pose. \cite{7759620} extend the steps of SVO \cite{6906584} to handle lines. In this case, the intensity residual for a given line is defined as the photometric error between sampled pixels on the 3D line. Notably, the usage of point and line features does not exhibit significant differences in these scenarios.

\begin{figure}[ht]
  \centering
  \includegraphics[width=0.4\textwidth]{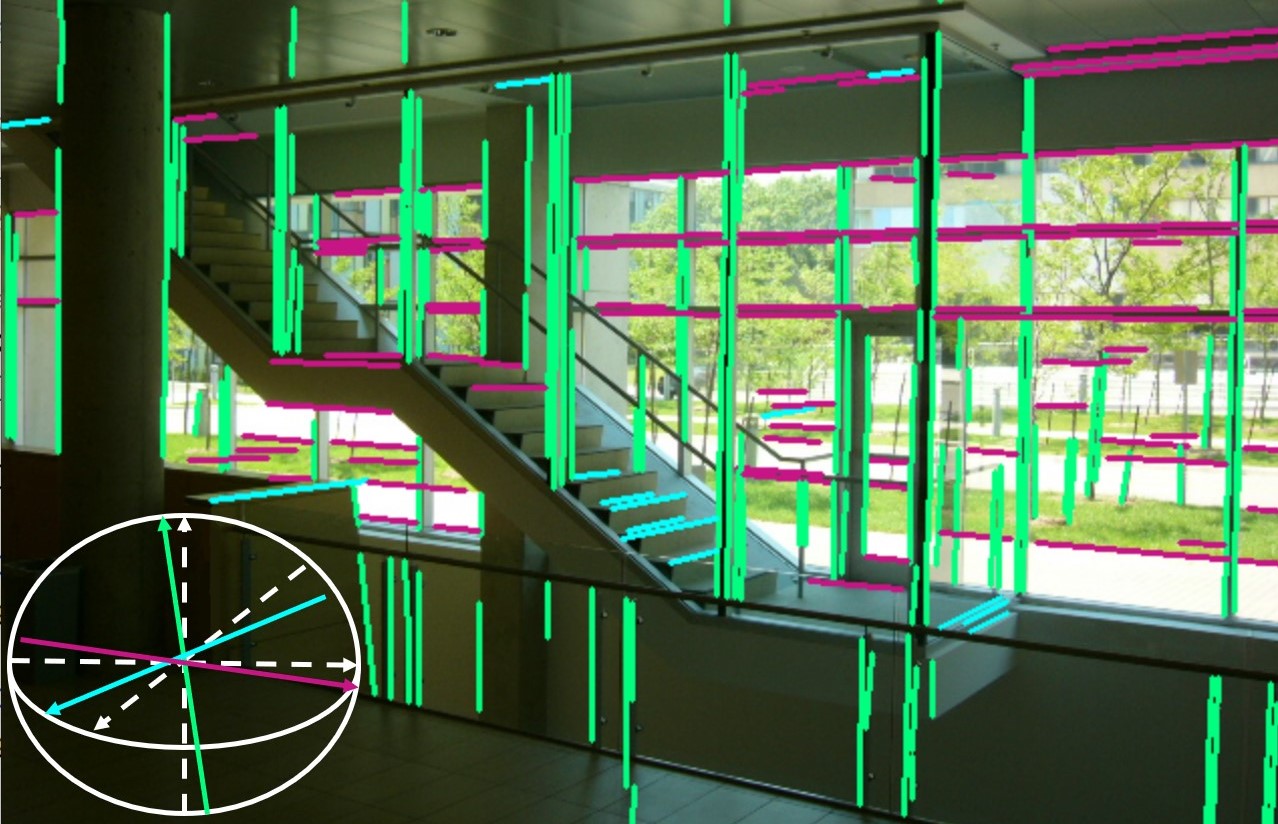}
  \caption{Example of typical structural scene. Lines in the scene have parallel of perpendicular relationship with the axis of MA, which can be used for optimization.}
  \label{MA.jpg}
\end{figure}

To leverage line features, several algorithms \cite{li2021rgb, li2020structure, kim2018low, company2022msc, straub2014mixture, yunus2021manhattanslam} incorporate the Manhattan world hypothesis into their frameworks. The Manhattan world hypothesis \cite{coughlan1999manhattan} asserts that in structured scenes like Fig. \ref{MA.jpg}, it is possible to extract a perpendicular Manhattan Axis (MA). Lines in these scenes exhibit parallel or vertical relationships with the axes of MA. However, the calculation of MA poses a significant challenge in its utilization. Current methods predominantly rely on plane normal vectors and line direction vectors to compute MA. But extraction of plane is more challenging than point and line. \cite{li2020structure} applies neural network to segment plane regions and estimate plane normal vectors. \cite{holz2012real} \cite{kim2018low} utilize integral graph of pixel normal vector to extract parameter of plane. Given the real-time requirements of visual odometry, the development of a fast and precise method for detecting normal vectors becomes paramount for the successful integration of the MA.

\section{METHOD}
\label{METHOD}

\begin{figure}[ht]
  \centering
  \includegraphics[width=0.48\textwidth]{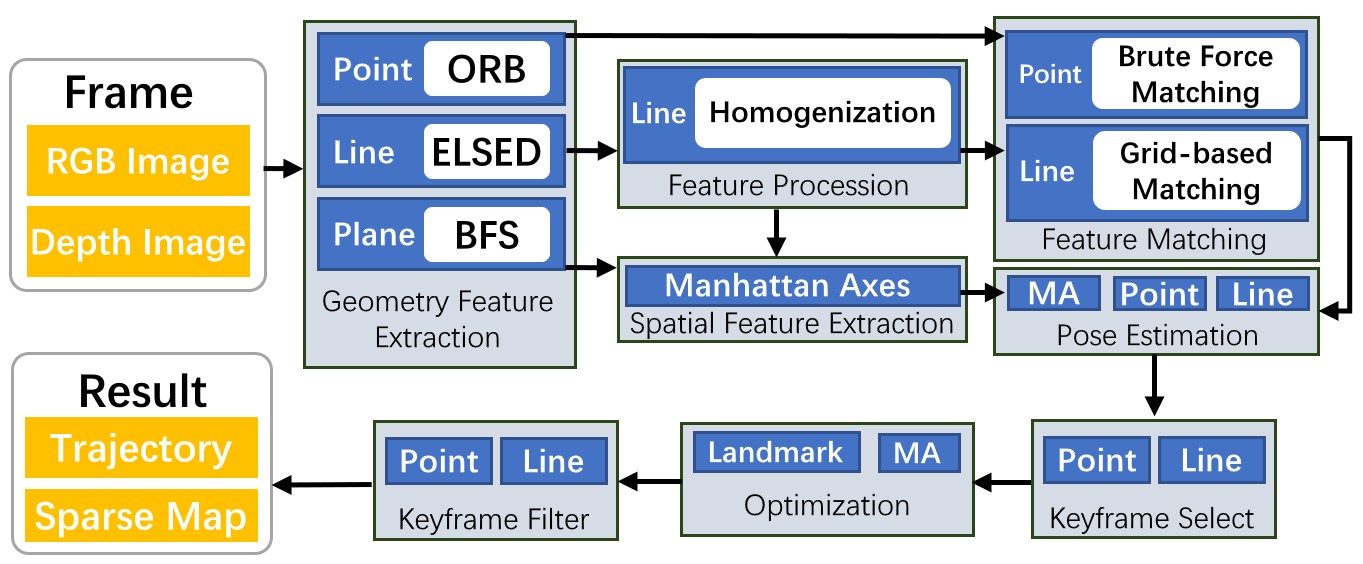}
  \caption{Overview of GFS-VO}
  \label{Overview of GFS-VO}
\end{figure}

The structure of GFS-VO is demonstrated in Fig. \ref{Overview of GFS-VO}. The system start with geometry feature extraction. In spatical feature extraction, we use homogenized lines and plane normal vector to calculate MA. Multi feature constraint will be used in the following pose estimation and optimization. Further details can be found in the following of this chapter.

\subsection{Feature Extraction}

% \subsubsection{Geometry Feature}

Feature used in GFS-VO can be divided in two kinds: geometry and spatial. Geometry features are extracted using separate threads. For point features, we utilize ORB\cite{rublee2011orb} to extract and describe points. In contrast to commonly used method \cite{von2012lsd}, we employ EDLine\cite{akinlar2011edlines} for line detection. Considering long lines provide more stable observations across frames, we adopt line connection strategy proposed in \cite{suarez2022elsed}. This strategy extends broken lines along the current gradient direction to connect them. Finally, extracted lines are described using LBD descriptor \cite{zhang2013efficient}.
\begin{figure}[ht]
  \centering
  \includegraphics[width=0.4\textwidth]{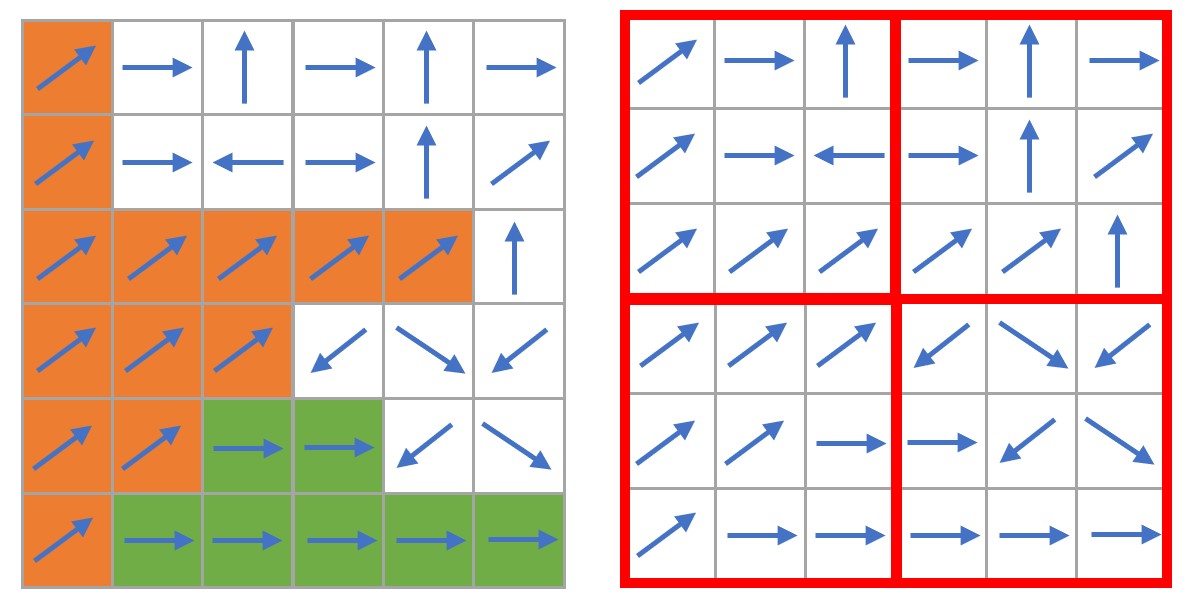}
  \caption{The comparison of plane normal extraction algorithm. Left is the result of our BFS-based algorithm while right is integral graph based algorithm. Compared with traditional method, our method can break the limit of regular grid and less affected by noise.}   
  \label{2.jpg}
\end{figure}

In GFS-VO, we incorporate plane feature primarily to extract MA, which relies on plane normal vectors. To achieve accurate and efficient extraction, we devise a Breadth-First Search based approach. Algorithm \ref{BFS} provides an overview of our method. Firstly, we reconstruct 3D positions of pixels using depth image. Next, we examine the angle between normal vectors of current pixel and adjacent pixels. If the angle is below threshold, we consider these two pixels to have the same direction and belong to a same plane. Subsequently, we perform successive search to identify and count the number of pixels with same direction within one search. Only planes with enough same direction pixels are deemed valid planes. To determine the normal vector of a plane, we compute the average of normal vectors of all same direction pixels associated with that plane. Fig .\ref{2.jpg} illustrates the distinction between our method and other approaches.
\begin{algorithm}[h]  
  \caption{BFS Based Normal vector Extraction}  
  \label{BFS}  
  \begin{algorithmic}[1]  
    \Require Rebuild depth image: $M$.
    \Ensure Plane normal vectors: $pt\_normal$.
    \Function{PixelNormalExtraction}{$M$, $P$}
    \State  $\vec{left},\vec{right},\vec{up},\vec{down}$=$getAdjacent($M$,$P$)$;
    \State  \Return{$(\vec{right} - \vec{left})\times(\vec{up} - \vec{down})$};
    \EndFunction

    \Function{BFS}{$pixel, M$}
    \State queue $q$ = initQueue($pixel$);
    \While{!$q$.empty()}
      \State pushOutFront($n$, $q$)
      \If {!$hasCalEd(getAdjacent(n))$} 
        \State $PixelNormalExtraction(M, n)$;
      \EndIf
      \State $q$.insert($checkAround(M, n)$);
    \EndWhile
    \EndFunction

    \Function{NormalExtraction}{$M$}
      \For{$pixel$ in $M$}
        \If {!$hasPassed(pixel)$} 
          \State $pt\_normal$.add($BFS(pixel, M)$);
        \EndIf
      \EndFor
      \State \Return{$pt\_normal$};
    \EndFunction 
  \end{algorithmic}  
\end{algorithm}  
% \begin{figure}[ht]
%   \centering
%   \includegraphics[width=0.48\textwidth]{figures/3.jpg}
%   \caption{The result of planes extraction. We use same color to represent pixels from same plane. Only pixels of same plane will be used to calculate average normal vector.}
%   \label{2.jpg}
% \end{figure}
% \subsubsection{Spatial Feature}

Spatial feature used in our algorithm is mainly MA. We adopt method inspired by \cite{kim2018low}, which utilizes plane normal vectors and 3D line direction vectors. To get accurate 3D lines, after homogenization, we filter out depth-illegal pixels and offset-illegal pixels and then calculate parameters of 3d lines. Extracted spatial features will be employed in optimization which is introduced in section \ref{Visual Odometry}.

\subsection{Grid based Line Homogenization}

\subsubsection{Grid Structure}

We employ grid structure to divide image into separate areas, with each area referred to as a grid. The grid structure offers advantage of showing feature’s distribution in image. We then build a bipartite index to establish connections between grids and lines, which serves as the foundation of subsequent processes such as line homogenization and tracking.

\subsubsection{Line Homogenization Strategies}

As mentioned in chapter \ref{RELATED WORK}, the challenge in line homogenization primarily lies in node allocation. To address this issue, we propose three line homogenization methods. The idea is as follows:

\begin{itemize}
  \item \textbf{Quadtree based scheme}: For lines in the image, a marker is added to all the grids they traverse. The sum of markers within a grid is considered as record. 
  \item \textbf{Midpoint-Quadtree based scheme}: Lines are assigned to a specific grid based on the position of their midpoint. The sum of midpoints within a grid is record.
  \item \textbf{Score based scheme}: Linse within each grid are rewarded or penalized based on their average gradient. A scoring mechanism is employed to rank all lines, and a portion of lines with the highest scores are retained.
\end{itemize}

The first two scheme are extension of point homogenization. Our primary focus is on finding a unique node to represent a given line. In Quadtree-based scheme, we assign a record to every grid that line passes through. On the other hand, Midpoint-Quadtree scheme only adds a record to grid where the midpoint of line is located. As a result, the record of each grid can effectively describe density of lines within a specific range. Similar to point homogenization, subsequent division step can be performed based on records of grids.

\begin{figure}[ht]
  \centering
  \includegraphics[width=0.38\textwidth]{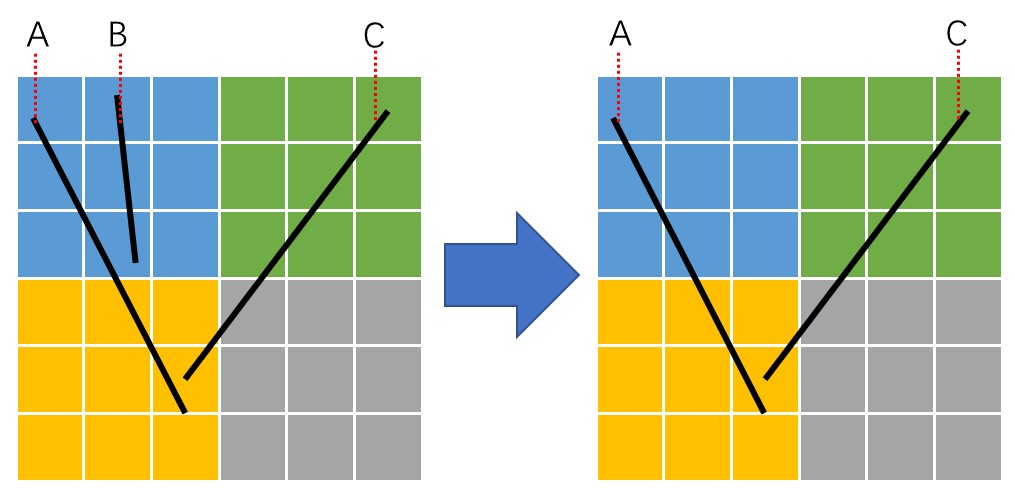}
  \caption{The example of dense part. We use four color to represent nodes in quadtree during division. In the orange grid, algorithm choose the most representative line $A$ and filter line $C$. But in green grid that $A$ doesn’t passed, the most representative line become $C$. So algorithm choose line $C$ to retain and cause incomplete homogenization in orange grid.}
  \label{4.jpg}
\end{figure}

Score-based scheme is designed from a global perspective. During the comparison experiment, we noticed that in some dense part of image like Fig. \ref*{4.jpg}, complete divide a line into corresponding region of a node becomes challenging due to line's inherent extension characteristics. Consequently, this often results in the incomplete homogenization. The key is that the selection of a line within a node does not imply that it is the best line, but rather that it is comparatively better than the other within a specific area. Building upon this concept, we replace division step in Quadtree or Midpoint-Quadtree based schemes with a scoring mechanism. In this approach, we prioritize lines with the highest average gradient and penalize the other within same grid. These rewards and punishments are reflected in the scores assigned to each line, which are determined based on the number of lines present in adjacent area of the grid.

Given that only one line in each grid is awarded while all remaining lines are punished, there is a significant disparity between numbers of punishment and award. To ensure fair selection of punished lines in the global screening process, we introduce an asymmetric score value, where the deduction for punished lines is less than the award. This asymmetric score value can be calculated as formula (1) and (2):
\begin{eqnarray}
  score & = & score + pow\left(na,4\right) \\
  score & = & score - pow\left(na,4\right)/ 2 \\
  score & = & score - pow\left(na,4\right)/ exp\left(np - 3\right)
\end{eqnarray}
where $na$ is the number of lines in adjacent area of current grid and $np$ is the number of grids that line has passed through. After calculating score of all lines, we retain lines with higher scores as the outcome of homogenization.

Furthermore, it is important to note that the average gradient can’t reflect line’s length. Consequently, longer lines tend to receive lower scores due to their participation in scoring multiple times. In scenarios where there is a significant disparity in line’s length, it is recommended to utilize formula (3), which reduces deduction value for longer lines by incorporating an exponential function in denominator.

% The whole process are summarized in Algorithm \ref{Score based line distribution}.

% \begin{algorithm}[ht]
%   \caption{Score-based line homogenization}
%   \label{Score based line distribution}
%   \begin{algorithmic}
%     \Require
%     Grid of lines: $Grid$. Extracted lines: $Keylines$. Threshold of distribution: $T$
%     \Ensure
%     Lines after distribution: $Resline$.
 
%     \State $Resline \gets \varnothing$
 
%     \For {$i \gets 0$ to $Grid.width$}
%           \For {$j \gets 0$ to $Grid.height$}          %For循环
%                   \State $lineG = Gridwindow(i, j)$ \Comment{lines in local grid}
%                   \State $lineSG = Gridwindow(i, j)$ \\\Comment{lines in surrounding grid}
%                   \State $tLine = selRep(lineG, Keylines)$ \\\Comment{Line with maximum mean gradient in local grid}
%                   \State $tLine.score += pow(lineSG, 4)$
%                   \For{$k \gets 0$ to $lineG.size()$}
%                   \State $teLine = Keylines[lineG[k]]$
%                   \If {$teLine != tLine$} 
%                   \State $teLine.score -= pow(lineSG, 4)/$\\$exp(lineG.size()-3)$
%                   \EndIf
%                   \EndFor
%           \EndFor
%     \EndFor
%     \State $sort(Keylines)$ \Comment{Rank according to score}
%     \For {$i \gets 0$ to $T$}
%     \State $Resline.push(Keylines[i])$
%     \EndFor 
%   \end{algorithmic}
% \end{algorithm}

\subsection{Visual Odometry}
\label{Visual Odometry}

\subsubsection{Grid-based Tracking}

When system is able to accurately estimate speed, the change between two frames is not expected to be significant. Under these circumstances matching time can be greatly reduced by leveraging grid structure. Specifically, we first find grids that the line passes through, as well as neighboring grids. Lines that pass through these grids are selected as candidate matches. Subsequently, descriptor matching is performed between given line and candidate lines. This approach reduces the number of candidate matches compared to one-to-one exhaustive calculation. Furthermore, it incorporates geometric positioning into the matching process, thereby enhancing both the accuracy and speed of matching procedure.
\begin{figure}[ht]
  \centering
  \includegraphics[width=0.4\textwidth]{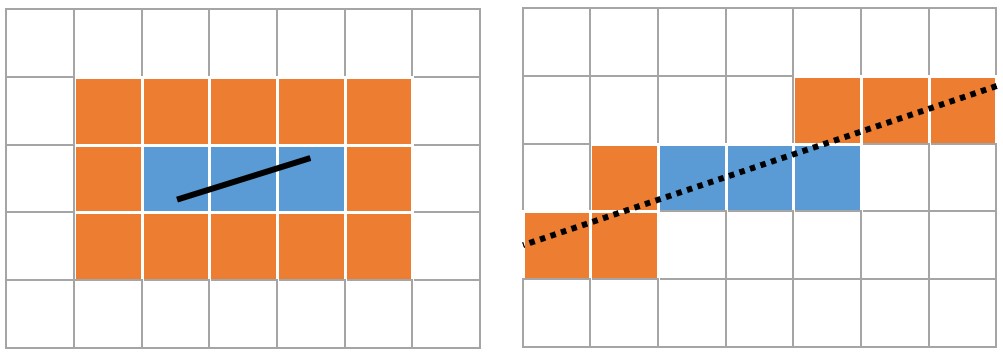}
  \caption{The illustration of search score expansion. Instead of searching in the surrounding grid, grids passed by line extension will also be searched.}
  \label{6.jpg}
\end{figure}
% \begin{eqnarray}
%   Candidate_{i} & = & lines \in lineWindow_{i}\\
%   lineWindow_{i} & = & Gridwindow\left(linePass_{i}\right)\\
% linePass_{i} & = & \left \{(x,y) |  line_{i} \cap Grid(x, y) \neq \varnothing \right \}  
% \end{eqnarray}
% where $Gridwindow$ is a function that returns the grid within adjacent area. 

In scenarios where estimated speed is unstable, matched line obtained from tracking tends to decrease. To address this issue, we can employ method of expanding search scope. Simply expanding the search radius, as done with point features, may not be effective due to the instability of line length. Nonetheless, the fracture of line segments does not alter their slope. Therefore, when tracking performance is poor, we can utilize an extension tracking method, as depicted in Fig. \ref{6.jpg}. The grids that the extension line passes through are then considered as an additional search range.
% It is worth noting that the grid-based tracking approach is particularly suitable for scenarios where the translation between two frames is not excessively large. However, in modules such as reference keyframe tracking, candidate matches from adjacent grids may exhibit inaccuracies. In such cases, we recommend employing global descriptor matching to achieve a more precise matching result.

\subsubsection{Pose Estimation}
Pose will be optimized when we get enough matched feature. For matched points and lines, we use constraints in \cite{mur2015orb} and \cite{pumarola2017pl} to calculate rotation and translation. For two consecutive frames, lowercase letters are used to represent features in the image plane, while uppercase letters represent features in world coordinates.
\begin{eqnarray}
  E_{P}(i,j) = \tau (i,j) \cdot \rho (\parallel p_{i} - \pi \left(P_{i},T\right) \parallel ^{2} )
  % E(P_{i},P_{j}) & = & \tau \left(P_{i}, P_{j}\right)\cdot \rho \left ( \parallel P_{i} - \pi \left(P_{ij}^{w},T\right) \parallel \right )\\
  % E_{L}(i,j) = & \tau (i,j) \cdot \rho (\parallel n_{i}·\pi \left(S_{i},T\right),n_{i}·\pi \left(E_{i},T\right) \parallel ^{2} )
  % E(L_{i}, L_{j}) & = & \tau \left(L_{i}, L_{j}\right)\cdot \rho \left (\right. \parallel n_{i}\cdot \pi \left(S_{ij}^{w},T\right)\nonumber \\
% &&,n_{i}\cdot \pi \left(E_{ij}^{w},T\right) \parallel \left. \right )
\end{eqnarray}
where $\tau$ is a binary function which return 1 only if point $i$ and $j$ is matched. $\rho$ is Huber loss function to reduce interfaces of noise. $\pi$ is projection function used to project 3d points into image plane. In a similar way, the reprojection error of a map line can be defined as:
\begin{eqnarray}
  E_{L}(i,j) = \tau (i,j) \cdot \rho (\parallel n_{i} \cdot \pi \left(S_{i},T\right),n_{i} \cdot \pi \left(E_{i},T\right) \parallel ^{2} )
  % E(L_{i}, L_{j}) & = & \tau \left(L_{i}, L_{j}\right)\cdot \rho \left (\right. \parallel n_{i}\cdot \pi \left(S_{ij}^{w},T\right)\nonumber \\
% &&,n_{i}\cdot \pi \left(E_{ij}^{w},T\right) \parallel \left. \right )
\end{eqnarray}
where $S_{i}$ and $E_{i}$ represent the coordinates of two endpoints of line $i$ and $n_{i}$ represents line's normal vector. Base on this, loss function needs to be minimized can be defined as: 
\begin{eqnarray}
    T = \underset{T}{\operatorname{argmin}} \left( \sum \limits_{\begin{subarray}{1}p_{i} \in p_{k}\\p_{j} \in p_{k - 1}\end{subarray}}^{}E_{P}(i,j) +\sum \limits_{\begin{subarray}{1}l_{i} \in l_{k}\\l_{j} \in l_{k - 1}\end{subarray}}^{}E_{L}(i,j)  \right)
\end{eqnarray}
Formula (9) will be optimized by LM algorithm in g2o library \cite{kummerle2011g}. Once the pose of the current frame is determined, we establish the connection between local map and current frame by utilizing both the pose and observation information.
        
\subsubsection{Keyframe Select and Filter}
Line homogenization introduces some instability to line features, which can result in reduction of map lines and reduction in their observation results. This may lead to tracking deviations. To address this issue, we propose two solutions. Firstly, we adjust the threshold for point and line observations when selecting keyframes, thereby weakening the connection between keyframes and local map. This adjustment helps mitigate the impact of line instability on the overall system. Secondly, in filtering keyframe, we extend point-based strategy to a point-and-line strategy. This means that a keyframe will only be considered redundant if there is a significant overlap in observations between both points and lines. By incorporating both point and line information, we ensure a more robust determination of redundant keyframes.

\subsubsection{Local Optimization}

Considering structural constraints existing in line segments and MA, the way used in \cite{company2022msc} is adopted here to embed structural constraints into optimization. The pose of covisible keyframe and coordinates of covisible elements will be optimizated. For a given keyframe $k$, we can get keyframes $K_c$ that have covisibility relationships from convisibility graph. We use $P$ and $L$ to represent map point and line seen by $K_c$. Set of keyframes $K_f$ that observe $P$ and $L$ but don’t connect to $K$ are also considered in optimization, but their pose is fixed. What we need to optimize is $\mathbb{N} = \left\{P_{i}^{w},L_{j}^{w},T_{k}\left|i \in P,j \in L,k \in K_{c}\right.\right\}$, so the loss function is:
\begin{equation}
    \mathbb{N}  =  \underset{\mathbb{N}}{\operatorname{argmin}}\left(\right. \sum \limits_{x \in K_{c} \cup K_{f}}^{}E_{R}^{x} + \sum \limits_{y \in K_{c}}^{}E_{S}^{y}  
    + \sum \limits_{z \in \mathbb{M}}^{}E_{M}^{z} \left. \right)
\end{equation}

This formula is composed of three parts, which respectively correspond to reprojection error($E_R$), structural constraints($E_S$), and parallel relation constraints with manhattan axis($E_M$). Specifically, $E_R$ is similar to the definition in formula (6), but match relationship changes to 2d feature and map elements. Structural constraints error $E_S$ is used to reflect the parallel and vertical relationship, which can be defined as:
\begin{equation}
  E_{S}^{y} = \sum \limits_{\left(i,j\right) \in L_{y}^{ \perp }}^{}\rho \left(E_{\left(i,j\right)}^{ \perp }\right) + \sum \limits_{\left(i,j\right) \in L_{y}^{ \parallel }}^{}\rho \left(E_{\left(i,j\right)}^{ \parallel }\right)
  \end{equation}
where $L_{k}^{ \perp }$ and $L_{k}^{ \parallel }$ are sets of perpendicular and parallel line pairs in $L$. $E^{ \perp }$ and $E^{ \parallel }$ is given by:
% \begin{eqnarray}
%   E_{\left(i,j\right)}^{ \perp } &  = &  \left|\cos(L_{i}^{c},L_{j}^{c})\right|\\
%   E_{\left(i,j\right)}^{ \parallel } &  = &  \sqrt{1 - \cos ^{2}(L_{i}^{c},L_{j}^{c})}
%   \end{eqnarray}
\begin{eqnarray}
  E_{\left(i,j\right)}^{ \perp } & = & \left|\cos(L_{i}^{c},L_{j}^{c})\right|\\
  E_{\left(i,j\right)}^{ \parallel } & = & \left|\sin(L_{i}^{c},L_{j}^{c})\right|
\end{eqnarray}

Finally, We use $\mathbb{M}$ to represent map line that associated with a MA and seen by any keyframe in $K_c$. $E_M$ is used to reflect the parallel relationship between given line and extracted MA, which can be given by:
\begin{equation}
E_{M}^{z}   =   E_{\left(z,Mz\right)}^{ \parallel }
\end{equation}

\section{EXPERIMENT RESULT}
\label{EXPERIMENT RESULT}

To check performance of our algorithm, we carried out sufficient experiments and compared with the latest algorithms. Considering that dataset collected in real scene always exist depth-illegal pixels, we also examine our performance in virtual scenes. All experiments have been performed on an Intel Core i5-10400 CPU @ 2.90GHz × 12/16GB RAM, without GPU parallelization. 

\subsection{Line homogenization}

\begin{figure}[ht]
  \centering
  \includegraphics[width=0.48\textwidth]{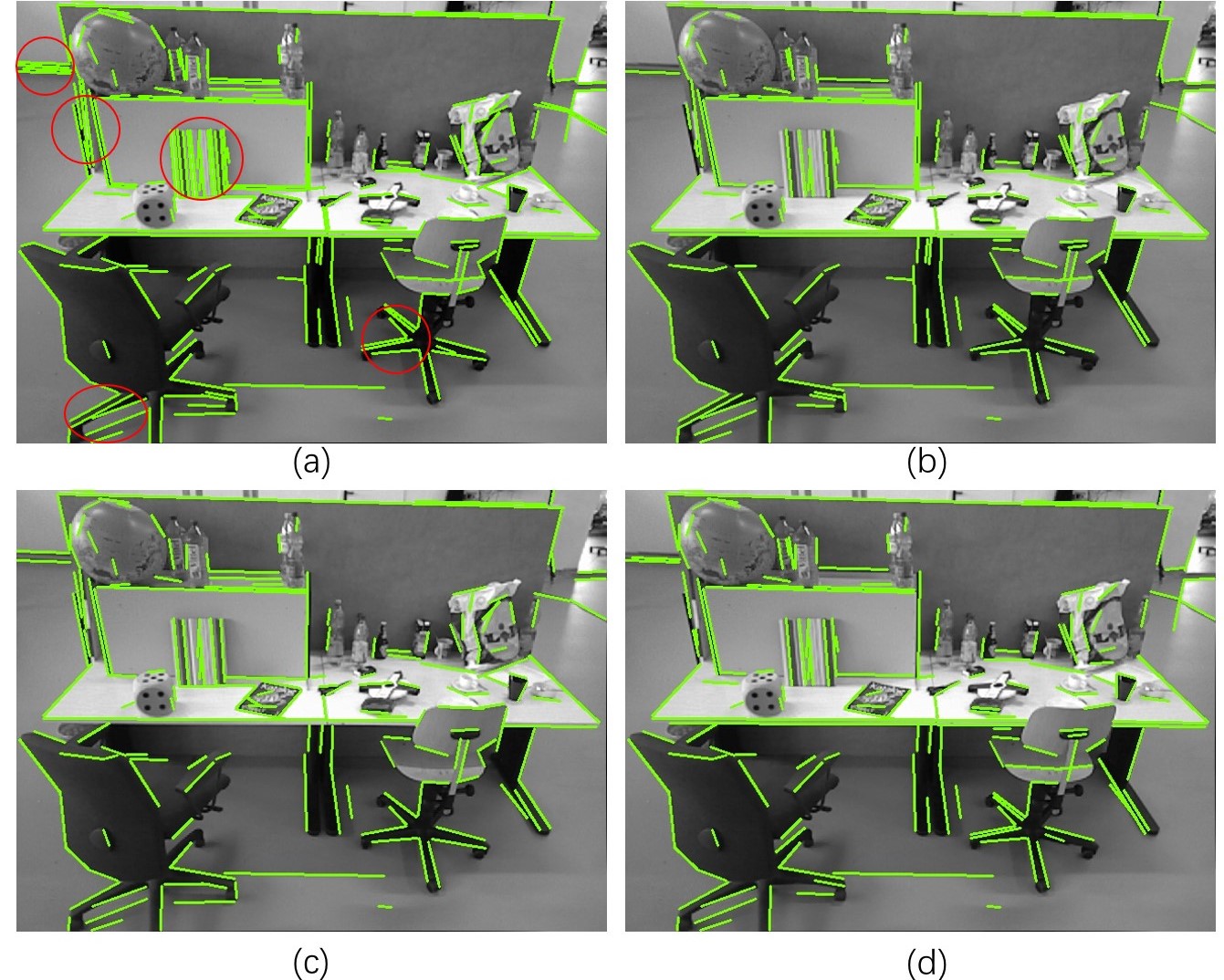}
  \caption{The result of proposed line homogenization algorithm.}
  \label{7.jpg}
\end{figure}

Fig. \ref{7.jpg} presents results of homogenization in a randomly selected image from TUM dataset. Dense areas in image are highlighted by red circles (Fig. \ref{7.jpg}(a)). It can be observed that each of the three methods has its own advantages.

Score-based scheme (Fig. \ref{7.jpg}(b)) exhibits noticeable performance, particularly in highly dense regions. This is primarily because, in such areas, the increase or decrease in score is largely influenced by the number of lines in adjacent grid. As a result, it becomes highly unlikely for poor-quality lines to be retained during subsequent screening process. Midpoint-Quadtree based scheme (Fig. \ref{7.jpg}(c)) represents the most balanced approach from a global perspective, as it selects only one line to be retained in each grid. This strategy ensures a more even distribution of retained lines throughout the map. However, the quadtree-based strategy (Fig. \ref{7.jpg}(d)) may face challenges in detecting dense locations, which leads to relatively weaker performance compared to the other two methods.

As SLAM is a framework with stringent real-time requirements, we also assess time consumption of proposed homogenization algorithm. We randomly select a group of images from TUM dataset and recorded their processing times. Our results show that the Score-based and Midpoint-Quadtree-based methods outperform the Quadtree-based method, with an average processing time of \textbf{4ms}. However, the Quadtree-based method still performs well, taking only 6ms to complete. Our findings indicate that there are no significant speed differences among the three methods. All three strategies effectively filter lines within dense areas of the image without significantly impacting overall processing speed.
% \begin{table}[ht]
%   \centering
%   \caption{Average Time Cost of Line homogenization Algorithm}
%   \label{Time Cost of three Line Uniformity Algorithm}
%   \resizebox{\linewidth}{!}{ 
%   \begin{tabular}{@{}|c|c|c|c|@{}} \hline
%             & Score & Midpoint-Quadtree & Quadtree  \\ \hline
%   Time cost & 0.0041s  & 0.0042s          & 0.0058s   \\ \hline
%   \end{tabular}}
% \end{table}

% It is important to highlight that the use of connection strategy in line extraction may result in smaller quantity of lines compared to the output of LSD. Furthermore, line homogenization further filters out lines in dense areas. Therefore, in sparsely distributed datasets, it becomes necessary to adjust the threshold for line extraction and homogenization to ensure an adequate number of lines are maintained.

\subsection{Framework Performance Comparison}

We select widely used RGB-D datasets \textbf{ICL-NUIM} \cite{sturm2012benchmark} and \textbf{TUM-RGBD} \cite{handa2014benchmark} to evaluate performance of our framework. ICL-NUIM consists of eight indoor sequences captured in two different scenes. These scenes present challenges such as low-textured regions and uneven feature distributions, which can lead to pose estimation deviations. Additionally, this dataset has optimized depth map noise, ensuring all pixel depths are valid. TUM also comprises several indoor sequences captured under various environmental conditions. Unlike ICL-NUIM, TUM-RGBD includes depth noise. We utilize ICL-NUIM to evaluate performance under ideal conditions and TUM-RGBD to assess performance in real-world scenes.

\subsubsection{Time Performance Comparison}

\begin{table}[ht]
  \centering
  \caption{Time Cost of Feature Extraction (In Second)}
  \label{Time Cost of Feature Extraction (In Second)}
  \resizebox{\linewidth}{!}{
  \begin{tabular}{|c|c|c|c|} \hline
    \multirow{2}{*}{Sequence} &\multirow{2}{*}{MSC-VO\cite{company2022msc}} &\multicolumn{2}{c|}{GFS-VO} \\ \cline{3-4}
    \multirow{2}{*}{}         &\multirow{2}{*}{}                            &Midpoint-Quadtree&Score \\ \hline
    fr1\_xyz          & 0.2833/0.0473 & 0.0972/0.0296           & 0.1550/0.0313 \\
    fr1\_desk         & 0.4225/0.0475 & 0.1315/0.0326           & 0.2111/0.0298 \\
    fr3\_longoffice   & 0.5434/0.0487 & 0.1735/0.0289           & 0.1289/0.0306 \\ \hline
    lr\_kt0            & 0.1291/0.0288 & 0.0641/0.0203           & 0.0657/0.0186 \\
    lr\_kt1            & 0.2157/0.0363 & 0.0197/0.0225           & 0.0231/0.0238 \\
    lr\_kt2            & 0.1777/0.0349 & 0.0347/0.0228           & 0.0352/0.0231 \\
    lr\_kt3            & 0.1604/0.0305 & 0.0319/0.0191           & 0.0318/0.0178 \\ \hline
    of\_kt0            & 0.1359/0.0361 & 0.0627/0.0250           & 0.0449/0.0246 \\
    of\_kt1            & 0.1657/0.0327 & 0.0426/0.0191           & 0.0546/0.0286 \\
    of\_kt2            & 0.1513/0.0343 & 0.0399/0.0225           & 0.0774/0.0227 \\
    of\_kt3            & 0.1247/0.0414 & 0.0468/0.0241           & 0.0321/0.0277 \\ \hline
    \end{tabular}
  }
\end{table}

To assess time cost of proposed feature extraction method, we conducted a comparison between GFS-VO and MSC-VO\cite{company2022msc}, both utilizing MA for trajectory estimation. The results, presented in Table \ref{Time Cost of Feature Extraction (In Second)}, are represented by "MA extraction/feature process". It is important to note that in MSC-VO, the feature processing time includes both extraction and reconstruction of geometry features and the extraction of plane normal vectors. On the other hand, in GFS-VO, this processing time also encompasses grid distribution and line homogenization. Based on the results presented in Table \ref{Time Cost of Feature Extraction (In Second)}, it is evident that GFS-VO significantly improves the speed of feature extraction.
  
The reduction in time cost can be attributed to three main factors. Firstly, in geometry feature extraction, we employ EDLINE instead of LSD, which reduces line extraction time. Secondly, line homogenization scheme in our algorithm reduces the number of line segments involved in reconstruction. Lastly, BFS-based approach enables accurate and rapid extraction of normal vectors from image. This helps in reducing unstable features and mitigating the influence of noise during the MA extraction, thereby improving extraction speed without compromising accuracy.

\begin{figure}[ht]
  \centering
  \includegraphics[width=0.48\textwidth]{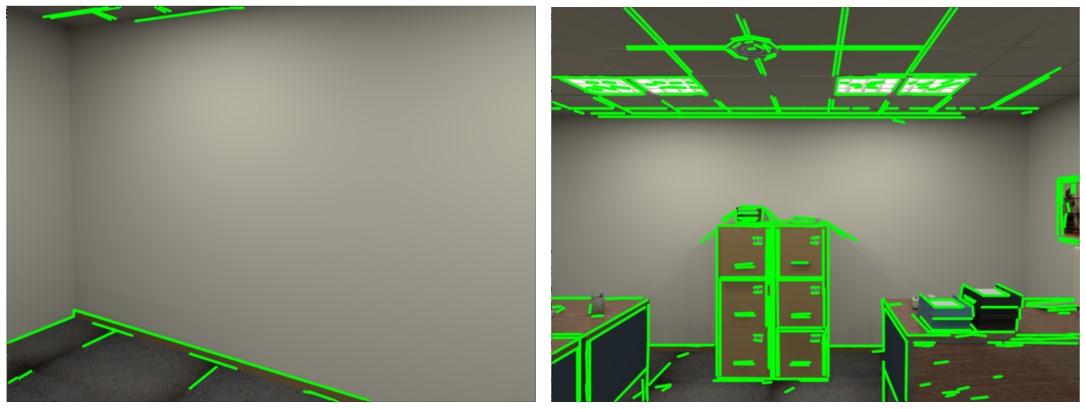}
  \caption{Example of scene with large change of feature amount. These images are from a same sequence of ICL-NUIM. The left image extracted 20 lines while the right extracted 323 lines. Under these circumstances, the setting of line homogenization threshold is always difficult.}
  \label{amount.jpg}
\end{figure}

At the same time, we also noticed deficiencies of GFS-VO. For datasets with large changes in the amount of line (like Fig. \ref{amount.jpg}), careful consideration should be given to setting the homogenization threshold. The main challenge arises from the fact that the threshold required for locations with abundant line features differs from that needed for sparser areas. Setting a small threshold effectively filters lines in rich locations, but it may not apply in sparse areas, and vice versa. There is no universal scale to measure the intensity of line features in all scenes.
\begin{table*}[ht]
  \centering
  \caption{RMSE of ATE for GFS-VO and other State-of-the-Art Framework (In Meters)}
  \label{RMSE of the ATE for GFS-VO and other State-of-the-Art Framework (In Meters)}
  \resizebox{\textwidth}{!}{ 
  \begin{tabular}{|c|c|c|c|c|c|c|c|c|c|c|} \hline
    \multirow{2}{*}{Sequence} &\multicolumn{2}{c|}{GFS-VO} &\multirow{2}{*}{MSC-VO\cite{company2022msc}} &\multirow{2}{*}{SReg\cite{li2021rgb}} &\multirow{2}{*}{ManhattanSLAM\cite{yunus2021manhattanslam}} &\multirow{2}{*}{LPVO\cite{8463207}}     &\multirow{2}{*}{Structure-SLAM\cite{li2020structure}} &\multirow{2}{*}{ORBSLAM2\cite{mur2017orb}} &\multirow{2}{*}{PS-SLAM\cite{zhang2019point}} &\multirow{2}{*}{InfiniTAM\cite{prisacariu2017infinitam}} \\ \cline{2-3}
    \multirow{2}{*}{}         &Midpoint-quadtree&Score       &\multirow{2}{*}{}                            &\multirow{2}{*}{}                     &\multirow{2}{*}{}                                           &\multirow{2}{*}{}                       &\multirow{2}{*}{}                                     &\multirow{2}{*}{}                          &\multirow{2}{*}{}                             &\multirow{2}{*}{}                                        \\ \hline
    lr\_kt0 &
    0.0082 &
    {\color[HTML]{F8A102} \textbf{0.0063}} &
    {\color[HTML]{F8A102} \textbf{0.006}} &
    {\color[HTML]{F8A102} \textbf{0.006}} &
    {\color[HTML]{3166FF} \textbf{0.007}} &
    0.01 &
    NA &
    0.025 &
    0.016 &
    NA \\
  lr\_kt1 &
    0.0116 &
    0.0105 &
    0.010 &
    0.015 &
    0.011 &
    0.04 &
    0.016 &
    {\color[HTML]{3166FF} \textbf{0.008}} &
    0.018 &
    {\color[HTML]{F8A102} \textbf{0.006}} \\
  lr\_kt2 &
    {\color[HTML]{F8A102} \textbf{0.0083}} &
    0.0102 &
    {\color[HTML]{3166FF} \textbf{0.009}} &
    0.020 &
    0.015 &
    0.03 &
    0.045 &
    0.023 &
    0.017 &
    0.013 \\
  lr\_kt3 &
    0.0152 &
    0.0241 &
    0.038 &
    {\color[HTML]{3166FF} \textbf{0.012}} &
    {\color[HTML]{F8A102} \textbf{0.011}} &
    0.10 &
    0.046 &
    0.021 &
    0.025 &
    NA \\ \hline
  of\_kt0 &
    {\color[HTML]{3166FF} \textbf{0.0210}} &
    {\color[HTML]{F8A102} \textbf{0.0190}} &
    0.028 &
    0.041 &
    0.025 &
    0.06 &
    NA &
    0.037 &
    0.032 &
    0.042 \\
  of\_kt1 &
    {\color[HTML]{3166FF} \textbf{0.0160}} &
    0.0171 &
    0.017 &
    0.020 &
    {\color[HTML]{F8A102} \textbf{0.013}} &
    0.05 &
    NA &
    0.029 &
    0.019 &
    0.025 \\
  of\_kt2 &
    0.0146 &
    {\color[HTML]{3166FF} \textbf{0.0127}} &
    0.014 &
    {\color[HTML]{F8A102} \textbf{0.011}} &
    0.015 &
    0.04 &
    0.031 &
    0.039 &
    0.026 &
    NA \\
  of\_kt3 &
    {\color[HTML]{3166FF} \textbf{0.0108}} &
    {\color[HTML]{F8A102} \textbf{0.0096}} &
    {\color[HTML]{3166FF} \textbf{0.010}} &
    0.014 &
    0.013 &
    0.03 &
    0.065 &
    0.065 &
    0.012 &
    {\color[HTML]{3166FF} \textbf{0.010}} \\ \hline
  fr1\_desk &
    0.0178 &
    {\color[HTML]{F8A102} \textbf{0.0167}} &
    0.019 &
    NA &
    0.027 &
    NA &
    NA &
    0.022 &
    0.026 &
    NA \\
  fr1\_xyz &
    {\color[HTML]{F8A102} \textbf{0.0094}} &
    {\color[HTML]{3166FF} \textbf{0.0109}} &
    0.010 &
    NA &
    0.010 &
    NA &
    NA &
    {\color[HTML]{3166FF} \textbf{0.010}} &
    0.010 &
    NA \\
  fr2\_desk &
    {\color[HTML]{F8A102} \textbf{0.0135}} &
    {\color[HTML]{3166FF} \textbf{0.0200}} &
    0.023 &
    NA &
    0.037 &
    NA &
    NA &
    0.040 &
    0.025 &
    NA \\
  fr2\_xyz &
    {\color[HTML]{F8A102} \textbf{0.0036}} &
    {\color[HTML]{3166FF} \textbf{0.0037}} &
    0.005 &
    NA &
    0.008 &
    NA &
    NA &
    0.009 &
    0.009 &
    NA \\
  fr3\_long\_office &
    {\color[HTML]{F8A102} \textbf{0.0188}} &
    {\color[HTML]{3166FF} \textbf{0.0209}} &
    0.022 &
    NA &
    NA &
    0.19 &
    NA &
    0.028 &
    NA &
    NA \\ \hline
  
  \end{tabular}
  }	
  \begin{tablenotes}
    \footnotesize
    \item * We use NA to stand for unavailable result. The best result is shown in orange while the second best is show in blue.  
  \end{tablenotes}
\end{table*}
\subsubsection{Accuracy Comparison}

We use Root-Mean-Square Error (RMSE) of absolute trajectory error as evaluation standard. The performance of other methods in experiment are from the best results provided in the respective papers. The comparison is shown in Tab. \ref{RMSE of the ATE for GFS-VO and other State-of-the-Art Framework (In Meters)}.

Upon experiment results, we notice that improvement in virtual scenes is limited. This limitation primarily stems from the scarcity of stable features in these datasets, particularly in the living room dataset. The instability of line features, in terms of length and quantity, not only affects the accuracy of pose estimation but also impacts MA extraction. Conversely, in TUM dataset, which represents real scenes, GFS-VO demonstrates significant improvement. We attribute this improvement to the complexity of point and line features in actual scenes, which highlights the benefits brought about by line homogenization. On one hand, our method preserves longer lines in the scene compared to traditional response-based line screening, establishing a more stable observation relationship between frames. On the other hand, homogenization removes short lines in dense areas. These short lines are more unstable and prone to errors in matching and pose estimation. The removal of such lines positively impacts the overall accuracy.

% It is worth mentioning that, from the perspective of a single image, quadtree-based strategy is less effective than score-based strategy in eliminating dense line segments. However, in terms of trajectory accuracy, there is no significant difference between two strategies. Both approaches significantly enhance the accuracy of pose estimation, especially when line segments in the dataset exhibit complexity.

\section{CONCLUSION}
\label{CONCLUSION}

This paper presents GFS-VO, a fast-structural visual odometer based on grid. Leveraging the grid structure, we design stable line homogenization and accurate line tracking algorithm. To fully use line feature, we introduce MA into our framework. Considering the real-time requirement of visual odometry, we also propose a plane normal vector extraction method to calculate MA faster. The experiment result shows that our method has a significant improvement in both accuracy and speed. For future work, we will continue to refine the line homogenization strategy and explore alternative approaches for measuring intensity. Furthermore, we also aim to investigate the impact of the position between point and line features on the accuracy of visual odometry, addressing the issues identified during our experiments.

\bibliographystyle{IEEEtran}
\bibliography{IEEEabrv,IEEEexample}

\begin{thebibliography}{10}
\providecommand{\url}[1]{#1}
\csname url@rmstyle\endcsname
\providecommand{\newblock}{\relax}
\providecommand{\bibinfo}[2]{#2}
\providecommand\BIBentrySTDinterwordspacing{\spaceskip=0pt\relax}
\providecommand\BIBentryALTinterwordstretchfactor{4}
\providecommand\BIBentryALTinterwordspacing{\spaceskip=\fontdimen2\font plus
\BIBentryALTinterwordstretchfactor\fontdimen3\font minus \fontdimen4\font\relax}
\providecommand\BIBforeignlanguage[2]{{%
\expandafter\ifx\csname l@#1\endcsname\relax
\typeout{** WARNING: IEEEtran.bst: No hyphenation pattern has been}%
\typeout{** loaded for the language `#1'. Using the pattern for}%
\typeout{** the default language instead.}%
\else
\language=\csname l@#1\endcsname
\fi
#2}}

\bibitem{von2012lsd}
R.~G. Von~Gioi, J.~Jakubowicz, J.-M. Morel, and G.~Randall, ``Lsd: A line segment detector,'' \emph{Image Processing On Line}, vol.~2, pp. 35--55, 2012.

\bibitem{mur2015orb}
R.~Mur-Artal, J.~M.~M. Montiel, and J.~D. Tardos, ``Orb-slam: a versatile and accurate monocular slam system,'' \emph{IEEE transactions on robotics}, vol.~31, no.~5, pp. 1147--1163, 2015.

\bibitem{mur2017orb}
R.~Mur-Artal and J.~D. Tard{\'o}s, ``Orb-slam2: An open-source slam system for monocular, stereo, and rgb-d cameras,'' \emph{IEEE transactions on robotics}, vol.~33, no.~5, pp. 1255--1262, 2017.

\bibitem{campos2021orb}
C.~Campos, R.~Elvira, J.~J.~G. Rodr{\'\i}guez, J.~M. Montiel, and J.~D. Tard{\'o}s, ``Orb-slam3: An accurate open-source library for visual, visual--inertial, and multimap slam,'' \emph{IEEE Transactions on Robotics}, vol.~37, no.~6, pp. 1874--1890, 2021.

\bibitem{qin2018vins}
T.~Qin, P.~Li, and S.~Shen, ``Vins-mono: A robust and versatile monocular visual-inertial state estimator,'' \emph{IEEE Transactions on Robotics}, vol.~34, no.~4, pp. 1004--1020, 2018.

\bibitem{pumarola2017pl}
A.~Pumarola, A.~Vakhitov, A.~Agudo, A.~Sanfeliu, and F.~Moreno-Noguer, ``Pl-slam: Real-time monocular visual slam with points and lines,'' in \emph{2017 IEEE international conference on robotics and automation (ICRA)}.\hskip 1em plus 0.5em minus 0.4em\relax IEEE, 2017, pp. 4503--4508.

\bibitem{fu2020pl}
Q.~Fu, J.~Wang, H.~Yu, I.~Ali, F.~Guo, Y.~He, and H.~Zhang, ``Pl-vins: Real-time monocular visual-inertial slam with point and line features,'' \emph{arXiv preprint arXiv:2009.07462}, 2020.

\bibitem{he2018pl}
Y.~He, J.~Zhao, Y.~Guo, W.~He, and K.~Yuan, ``Pl-vio: Tightly-coupled monocular visual--inertial odometry using point and line features,'' \emph{Sensors}, vol.~18, no.~4, p. 1159, 2018.

\bibitem{akinlar2011edlines}
C.~Akinlar and C.~Topal, ``Edlines: A real-time line segment detector with a false detection control,'' \emph{Pattern Recognition Letters}, vol.~32, no.~13, pp. 1633--1642, 2011.

\bibitem{suarez2022elsed}
I.~Su{\'a}rez, J.~M. Buenaposada, and L.~Baumela, ``Elsed: Enhanced line segment drawing,'' \emph{Pattern Recognition}, vol. 127, p. 108619, 2022.

\bibitem{zhou2022edplvo}
L.~Zhou, G.~Huang, Y.~Mao, S.~Wang, and M.~Kaess, ``Edplvo: Efficient direct point-line visual odometry,'' in \emph{2022 International Conference on Robotics and Automation (ICRA)}.\hskip 1em plus 0.5em minus 0.4em\relax IEEE, 2022, pp. 7559--7565.

\bibitem{9484792}
L.~Zhou, S.~Wang, and M.~Kaess, ``Dplvo: Direct point-line monocular visual odometry,'' \emph{IEEE Robotics and Automation Letters}, vol.~6, no.~4, pp. 7113--7120, 2021.

\bibitem{9393474}
Q.~Wang, Z.~Yan, J.~Wang, F.~Xue, W.~Ma, and H.~Zha, ``Line flow based simultaneous localization and mapping,'' \emph{IEEE Transactions on Robotics}, vol.~37, no.~5, pp. 1416--1432, 2021.

\bibitem{7759620}
R.~Gomez-Ojeda, J.~Briales, and J.~Gonzalez-Jimenez, ``Pl-svo: Semi-direct monocular visual odometry by combining points and line segments,'' in \emph{2016 IEEE/RSJ International Conference on Intelligent Robots and Systems (IROS)}, 2016, pp. 4211--4216.

\bibitem{6906584}
C.~Forster, M.~Pizzoli, and D.~Scaramuzza, ``Svo: Fast semi-direct monocular visual odometry,'' in \emph{2014 IEEE International Conference on Robotics and Automation (ICRA)}, 2014, pp. 15--22.

\bibitem{li2021rgb}
Y.~Li, R.~Yunus, N.~Brasch, N.~Navab, and F.~Tombari, ``Rgb-d slam with structural regularities,'' in \emph{2021 IEEE international conference on Robotics and automation (ICRA)}.\hskip 1em plus 0.5em minus 0.4em\relax IEEE, 2021, pp. 11\,581--11\,587.

\bibitem{li2020structure}
Y.~Li, N.~Brasch, Y.~Wang, N.~Navab, and F.~Tombari, ``Structure-slam: Low-drift monocular slam in indoor environments,'' \emph{IEEE Robotics and Automation Letters}, vol.~5, no.~4, pp. 6583--6590, 2020.

\bibitem{kim2018low}
P.~Kim, B.~Coltin, and H.~J. Kim, ``Low-drift visual odometry in structured environments by decoupling rotational and translational motion,'' in \emph{2018 IEEE international conference on Robotics and automation (ICRA)}.\hskip 1em plus 0.5em minus 0.4em\relax IEEE, 2018, pp. 7247--7253.

\bibitem{company2022msc}
J.~P. Company-Corcoles, E.~Garcia-Fidalgo, and A.~Ortiz, ``Msc-vo: Exploiting manhattan and structural constraints for visual odometry,'' \emph{IEEE Robotics and Automation Letters}, vol.~7, no.~2, pp. 2803--2810, 2022.

\bibitem{straub2014mixture}
J.~Straub, G.~Rosman, O.~Freifeld, J.~J. Leonard, and J.~W. Fisher, ``A mixture of manhattan frames: Beyond the manhattan world,'' in \emph{Proceedings of the IEEE Conference on Computer Vision and Pattern Recognition}, 2014, pp. 3770--3777.

\bibitem{yunus2021manhattanslam}
R.~Yunus, Y.~Li, and F.~Tombari, ``Manhattanslam: Robust planar tracking and mapping leveraging mixture of manhattan frames,'' in \emph{2021 IEEE International Conference on Robotics and Automation (ICRA)}.\hskip 1em plus 0.5em minus 0.4em\relax IEEE, 2021, pp. 6687--6693.

\bibitem{coughlan1999manhattan}
J.~M. Coughlan and A.~L. Yuille, ``Manhattan world: Compass direction from a single image by bayesian inference,'' in \emph{Proceedings of the seventh IEEE international conference on computer vision}, vol.~2.\hskip 1em plus 0.5em minus 0.4em\relax IEEE, 1999, pp. 941--947.

\bibitem{holz2012real}
D.~Holz, S.~Holzer, R.~B. Rusu, and S.~Behnke, ``Real-time plane segmentation using rgb-d cameras,'' in \emph{RoboCup 2011: Robot Soccer World Cup XV 15}.\hskip 1em plus 0.5em minus 0.4em\relax Springer, 2012, pp. 306--317.

\bibitem{rublee2011orb}
E.~Rublee, V.~Rabaud, K.~Konolige, and G.~Bradski, ``Orb: An efficient alternative to sift or surf,'' in \emph{2011 International conference on computer vision}.\hskip 1em plus 0.5em minus 0.4em\relax Ieee, 2011, pp. 2564--2571.

\bibitem{zhang2013efficient}
L.~Zhang and R.~Koch, ``An efficient and robust line segment matching approach based on lbd descriptor and pairwise geometric consistency,'' \emph{Journal of visual communication and image representation}, vol.~24, no.~7, pp. 794--805, 2013.

\bibitem{kummerle2011g}
R.~K{\""u}mmerle, G.~Grisetti, H.~Strasdat, K.~Konolige, and W.~Burgard, ``g 2 o: A general framework for graph optimization,'' in \emph{2011 IEEE International Conference on Robotics and Automation}.\hskip 1em plus 0.5em minus 0.4em\relax IEEE, 2011, pp. 3607--3613.

\bibitem{sturm2012benchmark}
J.~Sturm, N.~Engelhard, F.~Endres, W.~Burgard, and D.~Cremers, ``A benchmark for the evaluation of rgb-d slam systems,'' in \emph{2012 IEEE/RSJ international conference on intelligent robots and systems}.\hskip 1em plus 0.5em minus 0.4em\relax IEEE, 2012, pp. 573--580.

\bibitem{handa2014benchmark}
A.~Handa, T.~Whelan, J.~McDonald, and A.~J. Davison, ``A benchmark for rgb-d visual odometry, 3d reconstruction and slam,'' in \emph{2014 IEEE international conference on Robotics and automation (ICRA)}.\hskip 1em plus 0.5em minus 0.4em\relax IEEE, 2014, pp. 1524--1531.

\bibitem{8463207}
P.~Kim, B.~Coltin, and H.~J. Kim, ``Low-drift visual odometry in structured environments by decoupling rotational and translational motion,'' in \emph{2018 IEEE International Conference on Robotics and Automation (ICRA)}, 2018, pp. 7247--7253.

\bibitem{zhang2019point}
X.~Zhang, W.~Wang, X.~Qi, Z.~Liao, and R.~Wei, ``Point-plane slam using supposed planes for indoor environments,'' \emph{Sensors}, vol.~19, no.~17, p. 3795, 2019.

\bibitem{prisacariu2017infinitam}
V.~A. Prisacariu, O.~K{\""a}hler, S.~Golodetz, M.~Sapienza, T.~Cavallari, P.~H. Torr, and D.~W. Murray, ``Infinitam v3: A framework for large-scale 3d reconstruction with loop closure,'' \emph{arXiv preprint arXiv:1708.00783}, 2017.

\end{thebibliography}

% \end{thebibliography}

\end{document}